# Identifying safe intersection design through unsupervised feature extraction from satellite imagery

Jasper S. Wijnands[1] | Haifeng Zhao[1] | Kerry A. Nice[1] | Jason Thompson[1] | Katherine Scully[1] | Jingqiu Guo[2] | Mark Stevenson[1,3,4]

[1] Transport, Health and Urban Design Research Lab, Melbourne School of Design, The University of Melbourne, VIC, Australia

[2] Key Laboratory of Road and Traffic Engineering of the Ministry of Education, Tongji University, Shanghai, China

[3] Melbourne School of Engineering, The University of Melbourne, VIC, Australia

[4] Melbourne School of Population and Global Health, The University of Melbourne, VIC, Australia

**Correspondence**
Jasper S. Wijnands, Melbourne School of Design, The University of Melbourne, Masson Rd, Parkville VIC 3010, Australia.
Email: jasper.wijnands@unimelb.edu.au

**Funding information**
ACT Road Safety Fund, Grant/Award Number: 17/8281; LIEF HPC-GPGPU Facility, Grant/Award Number: LE170100200; Australian Research Council Discovery Early Career Researcher Award, Grant/Award Number: DE180101411; National Health and Medical Research Council, Grant/Award Number: APP1136250

**Abstract**

The World Health Organization has listed the design of safer intersections as a key intervention to reduce global road trauma. This article presents the first study to systematically analyze the design of all intersections in a large country, based on aerial imagery and deep learning. Approximately 900,000 satellite images were downloaded for all intersections in Australia and customized computer vision techniques emphasized the road infrastructure. A deep autoencoder extracted high-level features, including the intersection's type, size, shape, lane markings, and complexity, which were used to cluster similar designs. An Australian telematics data set linked infrastructure design to driving behaviors captured during 66 million kilometers of driving. This showed more frequent hard acceleration events (per vehicle) at four- than three-way intersections, relatively low hard deceleration frequencies at T-intersections, and consistently low average speeds on roundabouts. Overall, domain-specific feature extraction enabled the identification of infrastructure improvements that could result in safer driving behaviors, potentially reducing road trauma.

## 1 | INTRODUCTION

Each year, motor-vehicle crashes cause an estimated 1.35 million fatalities around the world and a further 50 million people incur nonfatal injuries (WHO, 2018). In fact, road deaths are the eighth leading cause of death globally and road traffic injuries are the number one cause of fatalities among individuals aged 5–29 years (WHO, 2018).

International bodies, such as the World Health Organization (WHO) and the United Nations (UN), have identified traffic safety as a pressing issue to be addressed. The UN has highlighted the need to reduce deaths and injuries caused by traffic accidents and to improve road safety as part of the Sustainable Development Goals 3 and 11 (UN, 2015). Furthermore, in recognizing the growing epidemic of road traffic deaths, the UN proclaimed the years 2011 to 2020 as a decade of action for road safety. The suggested strategic directions to reduce road trauma include improving road infrastructure and design, changing road user behavior, and enhancing vehicle safety. In particular, WHO argues that policymakers need to provide a greater focus on improving road design to reduce road injuries and deaths (WHO, 2018). For example, it has listed the design of safer intersections as a key intervention to improve road safety in its Save LIVES campaign (WHO, 2017).

The impact of intersection design on crash risk has been explored extensively (e.g., Björklund & Åberg, 2005; Young, Salmon, & Lenné, 2013). It has been recognized that the design of roads can be adapted to promote safe driving behavior, for example, through the introduction of roundabouts, signalized intersections, or speed humps (Martens, Comte, & Kaptein, 1997). Torok (2011) noted that altering road design to limit unsafe driving behavior is an easier and more time-effective solution to improving road safety than attempting to change driver behavior. In particular, Devlin, Candappa, Corben, and Logan (2011) argued that roads should be designed to minimize the occurrence or consequence of driver error. Such a concept is reflected in the "Safe System" approach adopted by many countries, focusing on creating safer roads, roadsides, and vehicles to accommodate human error when driving (OECD, 2008).

Intersections have been identified as crash "hot spots" for dangerous driving leading to crashes (Tay & Rifaat, 2007). According to Choi (2010), intersections are particularly hazardous due to activities such as turning across traffic, with the potential for obstructed views or inadequate surveillance while turning. Further, there are many possible points of conflict with other road users, including pedestrians, cyclists, and motorcyclists (Devlin et al., 2011). Transportation authorities such as the American Association of State Highway and Transportation Officials and the Transportation Association of Canada have provided detailed design guidelines for intersection configurations (TAC, 2017; AASHTO, 2018). These guidelines take into account traffic characteristics, physical elements, and human and economic factors. For example, safety can be affected by physical elements such as traffic islands, traffic lights, or the angle of the intersection. Many studies have reported that roundabouts offer the safest intersection type for road users (e.g., Rodegerdts et al., 2007).

### 1.1 | Current risk mitigation approaches

Efforts to reduce road trauma include the development of passive and active in-vehicle safety systems, the analysis of police-reported crash data to identify high-risk areas, driving simulator experiments, and site-specific case studies. For example, accident black spot initiatives have treated many intersections where crashes are common (e.g., BITRE, 2012). Traffic environments have been investigated using various approaches, including cameras placed at the intersection location (Sayed, Ismail, Zaki, & Autey, 2012; Thompson, Wijnands, Mavoa, Scully, & Stevenson, 2019), in-vehicle cameras (Dingus et al., 2006), and surrogate measures such as time-to-collision (Vogel, 2003). Further, Anwaar, Anastasopoulos, Ong, Labi, and Islam (2012) performed country-level analyses using summary statistics to inform road safety policies. The availability of detailed micro-level data at this large scale provides opportunities to further enhance road safety.

### 1.2 | Advances in data-driven methodologies

Recent advances in methodologies based on artificial intelligence (AI) allow for extensive analyses of large data sets (Schmidhuber, 2015). Contemporary methods include convolutional neural networks (CNNs) to analyze images (Szegedy, Vanhoucke, Ioffe, Shlens, & Wojna, 2016), and stacks of Long Short-Term Memory (LSTM) recurrent neural networks to process sequential data (Fernández, Graves, & Schmidhuber, 2007). Whereas these supervised learning methods require labeled data, unsupervised learning enables feature extraction in the absence of labels. In particular, by using the input data themselves as labels, autoencoders (AEs) can extract the main features by minimizing the difference between the input and reconstructed data, while passing the information through a narrow bottleneck layer (Ballard, 1987). Improvements in AE architectures, loss functions, and calibration approaches, such as partially corrupting input data, have further improved the robustness of extracted features (e.g., Vincent, Larochelle, Bengio, & Manzagol, 2008).

In the road safety domain, various applications have been proposed that incorporate these AI techniques. These applications include monitoring driving behavior using AEs (Guo, Liu, Zhang, & Wang, 2018) or LSTM recurrent neural networks (Wijnands, Thompson, Aschwanden, & Stevenson, 2018), drowsiness detection using CNNs (e.g., Park, Pan, Kang, & Yoo, 2017; Wijnands, Thompson, Nice, Aschwanden, & Stevenson, 2020), and style transfer using generative adversarial networks to identify safe infrastructure design (Zhao et al., 2019). All these deep learning implementations require very large data sets for model calibration. With respect to further optimizing intersection design, the following large-scale data sources provide particular opportunities: (i) in-vehicle telematics technology and (ii) satellite imagery.

### 1.3 | In-vehicle telematics technology

As crashes are rare events, small sample sizes complicate the analysis of causation with respect to road design factors (Devlin et al., 2011). Therefore, new large-scale data sources that capture driving behavior continuously could be valuable for exploring intersection safety, especially to identify near-miss events and minor crashes not recorded in police reports. In-vehicle telematics technology uses a smartphone application or a device installed into the on-board diagnostics port of a vehicle to record driving behavior from GPS information (Cassias & Kun, 2007). Some insurance companies use telematics devices to calculate personalized insurance premiums based on a driver's style, behavior, and ultimately, their crash risk (Handel et al., 2014). Its other uses include commercial fleet monitoring and performance monitoring for novice drivers (Horrey, Lesch, Dainoff, Robertson, & Noy, 2012; Stevenson et al., 2018).

### 1.4 | Satellite imagery

Large-scale data are also available for (relatively) static information, such as infrastructure design. In particular, satellite remote sensing provides detailed depictions of road infrastructure, globally. Most previous studies on this topic focus on extracting roads from satellite imagery. For example, Hu, Razdan, Femiani, Cui, and Wonka (2007) identified potential roads and intersections associated with each pixel in the image to grow a road network, which is pruned in a final step. Wang, Song, Chen, and Yang (2015) labeled satellite images with the road direction obtained from vector road maps and constructed the road network by tracking the predicted road direction in consecutive satellite images. Easa, Dong, and Li (2007) extracted curves after preprocessing imagery using edge detection and obtained the road network by connecting all extracted curves using straight lines. Further, Cadamuro, Muhebwa, and Taneja (2019) assessed road quality using a combination of AE and LSTM on raw satellite imagery.

Some studies have attempted to harness the information in satellite imagery for road safety analysis. For example, Najjar, Kaneko, and Miyanaga (2017) predicted city-scale road safety maps from raw satellite images. Further, Zhang, Lu, Zhang, Shang, and Wang (2018) used color and texture information and higher level features from pre-trained CNNs to identify locations with high crash risk. Both studies note that images at locations with elevated crash risk look similar, and distinctly different from the green colors and rooftops observed in locations with few crashes. However, this could merely be an indication of traffic volume and exposes a difficulty for interpreting raw satellite imagery in research of this kind; adding more trees to a complex intersection is unlikely to resolve a road safety issue. Therefore, what is required is the development of tailored methods to extract domain-specific features from raw satellite imagery for road safety analyses. By combining this with telematics data, our research proposes a deep learning approach to identify intersection design characteristics associated with (un)safe driving behaviors.

## 2 | METHODS

### 2.1 | Intersection identification

First, all intersection locations in Australia were identified using Python and OpenStreetMap (OSM). OSM provides open-source online maps generated by millions of volunteers. The complete map of Australia was downloaded from GeoFabrik (2018), providing free, daily updated maps. Osmosis (2018) was used to extract the road network of Australia as an OSM XML file. OSM uses tags to describe various attributes of map elements (e.g., "junction:roundabout", "highway:secondary"). Using these tags, only the subset of the road network accessible to motor-vehicles was selected.

The geometry is described in OSM using three elements: nodes, ways, and relations, where a node represents a geographic location, a way is a list of sequential nodes, and a relation is a collection of nodes and/or ways. As a node may appear in more than one way, intersections were identified as nodes that appeared in at least two ways. This resulted in a large collection of three-way intersections, including T and Y types, four-way intersections, multiway (i.e., more than four) intersections and nodes on roundabouts. The initial selection of intersections had to be refined further, for example, where there was no real-world opportunity for vehicles to turn (e.g., two-way streets splitting into two separate one-way roads). Other locations were incorrectly identified as intersections when the node was an endpoint of two different ways due to different street names or speed limits; these locations were also removed. Separated lanes sometimes led to multiple intersections being identified, although these points should be represented as a single intersection (see Figure 1). Therefore, for each set of identified nodes with small internode distances, the central location of the set was computed as the final intersection location. As this merging process is computationally expensive, a recursive partitioning algorithm was developed for this process to enable multicore processing.

A separate process was performed for roundabouts, where subsets of nodes were directly retrieved from the OSM XML file using the "junction:roundabout" attribute. For each roundabout, any separated segments were merged and the centroid was calculated based on all identified, nearby nodes. Checks were performed to remove (i) any incorrectly labeled segments and (ii) T-intersections of roads feeding into the roundabout. In total, a set of 903,704 geographic coordinates was obtained, corresponding to the centers of the intersections and roundabouts in Australia. The intersections were spread across the State of Victoria (26.8%), New South Wales (24.8%), Queensland (18.7%), Western Australia (13.3%), South Australia (10.4%), Tasmania (2.6%), Australian Capital Territory (2.0%), and Northern Territory (1.4%).

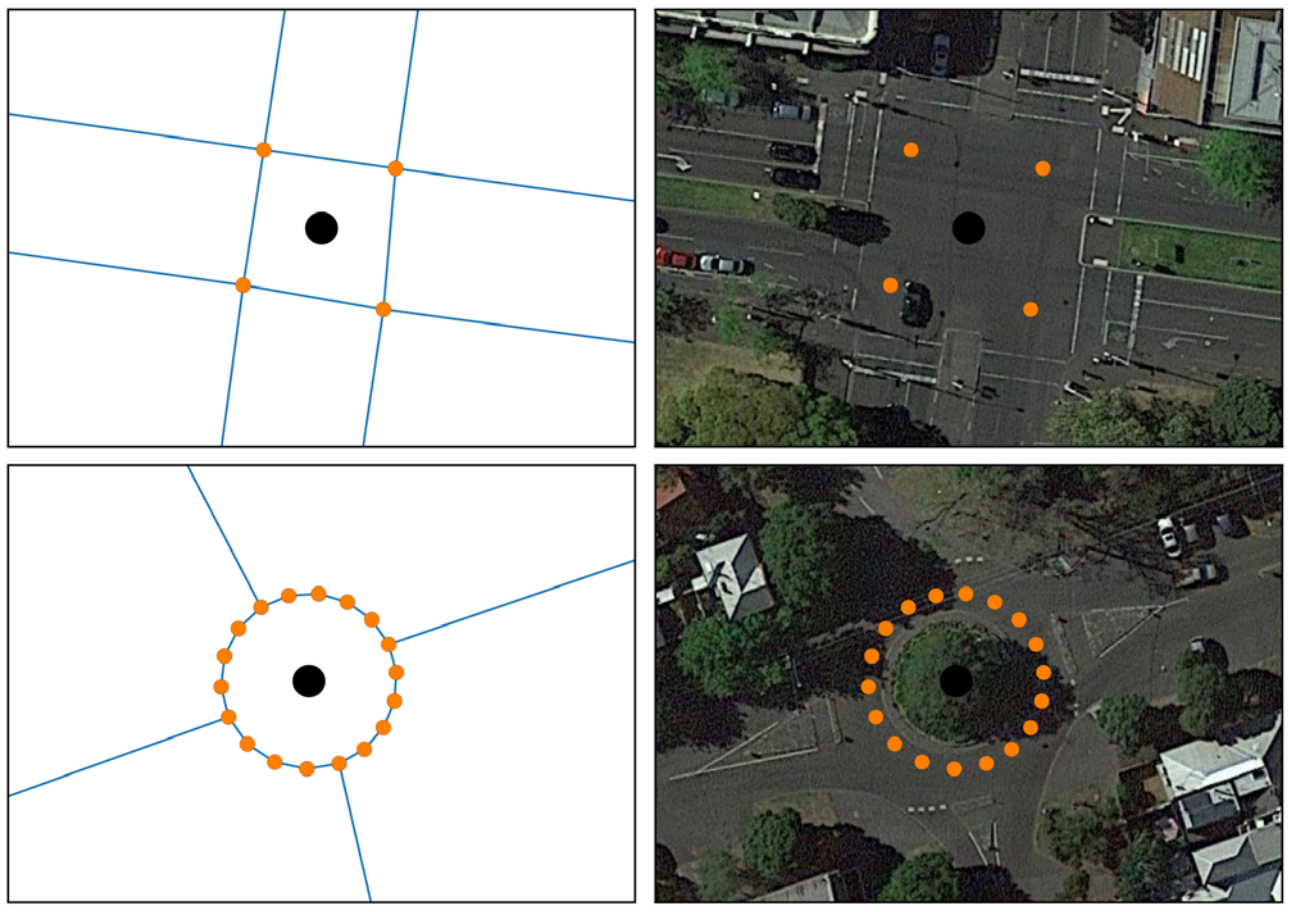

**FIGURE 1** Left panel: intersection identification using OSM geometry with initial steps (small, orange dots) and final location (large, black dot). Right panel: comparison to satellite imagery

## 2.2 | Obtaining and processing satellite imagery

Imagery based on satellite remote sensing was downloaded for all identified intersection locations using the Google Maps Static API. As images were not available at all locations, the sample size decreased to 898,418. The selected settings resulted in 256 × 256 pixels color images without labels at zoom level 19; two examples are provided in Figure 2a (Map data, 2018, Google). Not all information contained in these satellite images is important for road safety. Specifically, it would be relevant to capture the type, size, shape, lane markings, and complexity of an intersection. In contrast, the colors and shapes of rooftops, buildings, trees, and other vegetation are less important, especially when located away from the road near the edge of an image (i.e., not in the line of sight). To prevent clustering on these features, the images were preprocessed to emphasize road infrastructure and reduce the amount of redundant information. The creation of preprocessing steps was an iterative process where the impact on the quality of extracted features and image clustering was assessed through a validation process (see Sections 2.3–2.5). This led to the final set of preprocessing steps illustrated in Figure 2. First, using OpenCV (Bradski, 2000), images were (i) smoothed using Gaussian blur with a 3 × 3 kernel size, (ii) converted to grayscale, and (iii) processed using the Scharr operator (Jähne, Haussecker, & Geissler, 1999) to detect lines based on horizontal and vertical gradients (see Figure 2b). Since asphalt does not contain many lines or varying colors, as opposed to trees and buildings, these steps emphasize the road infrastructure.

Importantly, the location of the intersection is in the middle of the image by design (see Section 2.1). Therefore, it is possible to capture the same spatial range of information from the center of the intersection, regardless of image orientation. This was implemented using a fading scheme, preserving pixels within a 52-pixel radius from the center and gradually forcing pixel colors toward white at a radius of 128 pixels. This preprocessing step provided increased focus on intersection design and reduced elements such as shapes of buildings that the neural network could pick up (see Figure 2c).

An additional characteristic, which is still visible after these preprocessing steps, is the orientation of the road network. Since the heading of the road has only limited influence on road safety (e.g., adverse lighting conditions at sunrise/sunset could increase crash risk), it is best to prevent clustering based on the orientation of the intersection. As the visible image is circular, it can be rotated at any angle without information loss due to cropping. To identify the optimal rotation angle for an image, a discrete Fourier transform was applied, decomposing image $f(i, j)$ into its sine and cosine components (Equation (1)).

$$F(k,l) = \sum_{i=0}^{N-1} \sum_{j=0}^{N-1} f(i,j) e^{-\mathrm{i}2\pi\left(\frac{ki}{N}+\frac{lj}{N}\right)},$$

$$e^{\mathrm{i}x} = \cos x + \mathrm{i} \sin x \tag{1}$$

Figure 2d illustrates the magnitude component of $F$, showing high intensity perpendicular to the orientation of the main straight road in the image. The result was then transformed to a polar coordinate system where the mean intensity at each angle (i.e., per row in the image) was computed (see Figure 2e). This reduced the task of finding the rotation angle to determining the argmax value from 256 intensity values. For enhanced accuracy, a periodic cubic spline (Knott, 2000) was fitted to determine the rotation angle corresponding to the maximum intensity. Note that the order of magnitude of this refinement was around 0.5 degree. The image before application of the Fourier transform was then rotated by the computed angle to arrive at the final image (see Figure 2f). The resulting data set, containing 898,418 abstractions of all identified

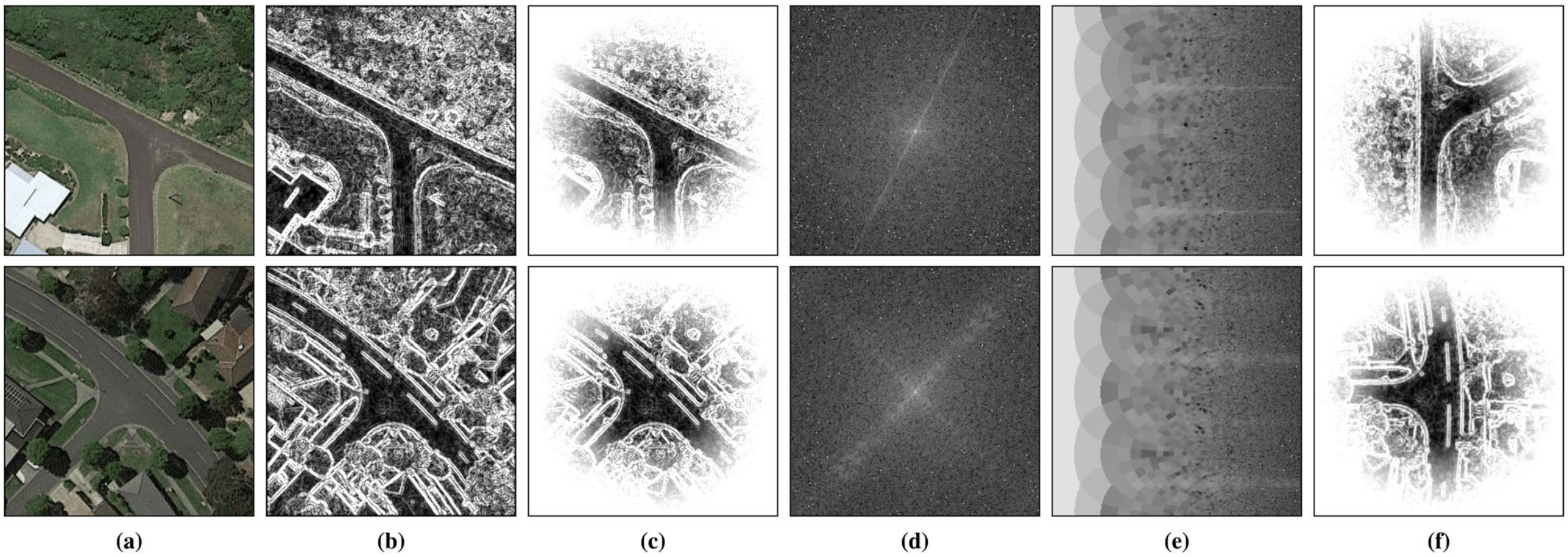

**FIGURE 2** Illustration of preprocessing steps for two sample images. The plots show (a) the original Google Street View image, (b) line detection based on the Scharr operator, (c) emphasizing intersection center, (d) Fourier transform, (e) transformation to log-polar space to determine rotation angle, and (f) the final rotated image

intersections in Australia, is publicly available through an open-access repository (Wijnands, 2020).

## 2.3 | Extracting features using a deep AE

The raw pixels in the generated data set are not directly suitable for accurate clustering. Therefore, a deep AE was developed using TensorFlow (Abadi et al., 2015) to extract the high-level features describing each image. AEs can be as simple as a three-layer network, consisting of an input, hidden, and output layer, with fully connected neurons (Rumelhart, Hinton, & Williams, 1985). However, for imagery, multiple successive layers with convolution filters have been shown to extract robust features (Long, Shelhamer, & Darrell, 2015). The network architecture designed for our study is described in Table 1. First, convolutional layers (Conv2D) encode the image by reducing the spatial dimensions, while increasing layer depth to extract increasingly complex features. This leads to a bottleneck layer $z$ of 2,048 neurons containing a condensed representation of the image. From here, transposed convolutional layers (Conv2D_T) decode the information and rebuild the $256 \times 256$ grayscale image. A $4 \times 4$ filter size was used in decoding layers to minimize checkerboard artifacts. Neurons in the AE use ReLU activation functions, except for the final layer. The model's robustness was further improved by batch normalization between convolutional layers.

The loss function $\mathcal{L}$ was formulated in this research as

$$\mathcal{L} = \sum_{i=1}^{256} \sum_{j=1}^{256} X_{ij} + \alpha \sum_{k} \|W_k\|_2 + \beta \|z\|_1 \qquad (2)$$

**TABLE 1** Description of autoencoder architecture

| Layer | Operation | Dimensions |
|---|---|---|
| 0 | Input image | 256 × 256 × 1 |
| 1 | Conv2D, 3×3 kernel, stride 2 | 128 × 128 × 64 |
| 2 | Conv2D, 3×3 kernel, stride 2 | 64 × 64 × 96 |
| 3 | Conv2D, 3×3 kernel, stride 2 | 32 × 32 × 128 |
| 4 | Conv2D, 3×3 kernel, stride 2 | 16 × 16 × 192 |
| 5 | Conv2D, 3×3 kernel, stride 2 | 8 × 8 × 256 |
| 6 | Conv2D, 3×3 kernel, stride 2 | 4 × 4 × 384 |
| 7 | Conv2D, 4×4 kernel, flatten | 2,048 |
| 8 | Conv2D_T, 4×4 kernel | 4 × 4 × 512 |
| 9 | Conv2D_T, 4×4 kernel, stride 2 | 8 × 8 × 512 |
| 10 | Conv2D_T, 4×4 kernel, stride 2 | 16 × 16 × 512 |
| 11 | Conv2D_T, 4×4 kernel, stride 2 | 32 × 32 × 384 |
| 12 | Conv2D_T, 4×4 kernel, stride 2 | 64 × 64 × 384 |
| 13 | Conv2D_T, 4×4 kernel, stride 2 | 128 × 128 × 256 |
| 14 | Conv2D_T, 4×4 kernel, stride 2 | 256 × 256 × 256 |
| 15 | Conv2D_T, 4×4 kernel, stride 1 | 256 × 256 × 1 |

with $X_{ij}$ the sigmoid cross-entropy loss for pixel $(i, j)$ and regularization terms including the L2 norm of all $k$ convolution kernel weight matrices ($W_k$) and the L1 norm of $z$'s activations. The latter is a sparsity constraint to improve feature quality by enforcing a sparse representation in the bottleneck layer. Hyperparameters $\alpha$ and $\beta$ control the relative importance of the regularization terms on overall model fit and were determined experimentally using grid search ($\alpha = 0.1$, $\beta = 0.05$). The AE was trained for 10 million iterations on an NVIDIA P100 graphics card (see Figure 3). Experiments with a further increased number of iterations did not lead to substantial improvements in AE output images.

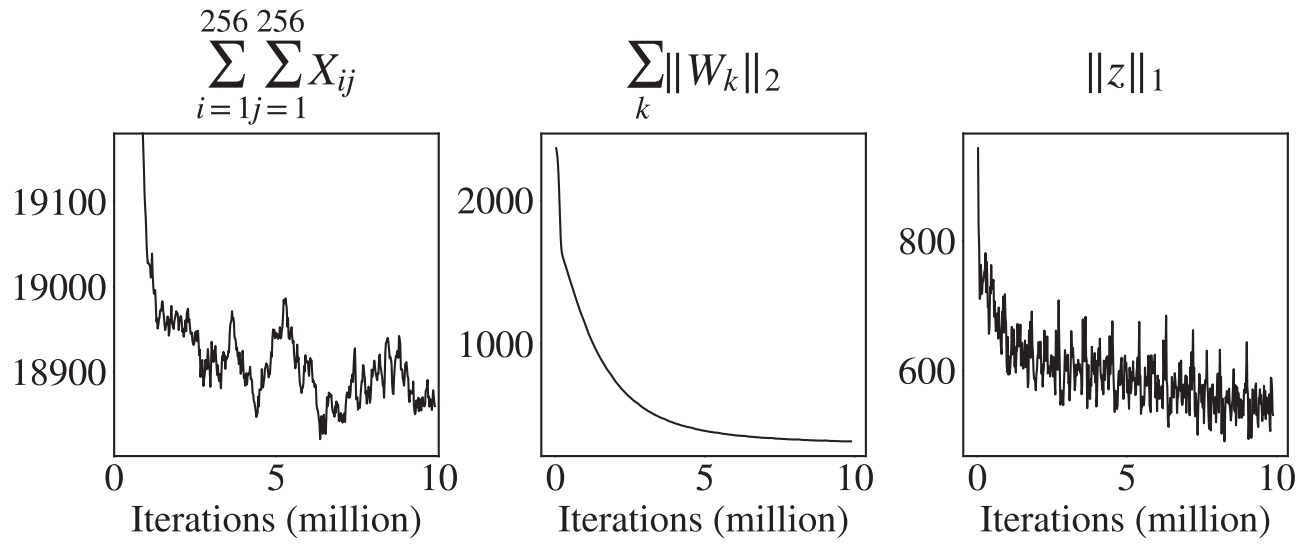

**FIGURE 3** Convergence of loss components

## 2.4 | Image clustering

All intersection abstractions were processed through the trained AE to obtain the 2,048 activations in $z$. This condensed representation was used to cluster images using the t-Distributed Stochastic Neighbor Embedding (t-SNE) algorithm (Van der Maaten & Hinton, 2008). t-SNE reduces the 2,048-dimensional input space to a 2-dimensional space, while aiming to preserve the relative distances between points in the high-dimensional space. The embedding for the $898{,}418 \times 2{,}048$ encoding was computed using the opt-SNE adaptation (Belkina et al., 2019), providing automatic optimization of several hyperparameters, based on a parallel t-SNE implementation by Ulyanov (2016). The perplexity of the t-SNE algorithm was set to 30 (recommended range [5, 50], see Van der Maaten and Hinton (2008)).

## 2.5 | Model validation

For validation purposes, hold-out OSM information was assigned to each intersection in the t-SNE embedding. Specifically, during the intersection identification process (see Section 2.1) it was identified what the number of legs of an intersection was and whether nodes had to be merged. This led to four "simplified" classes of intersections based on OSM classifications, splitting all intersections into the following basic categories: roundabouts (O), three-way intersections (T), simple four-way intersections (X), and complex intersections (#). The latter includes four-way intersections with at least one multi-lane leg and intersections with features such as dividing islands, slip lanes, or more than four legs. Further, traffic light information was obtained directly from OSM. As the simplified intersection class and traffic light information were not used in the methodology described above, they can provide an indication of the quality of the t-SNE clustering.

## 2.6 | Telematics data

An in-vehicle telematics data set from Australian insurer QBE/Insurance Box was used to obtain detailed measurements of driving behavior. The data set contained 272 million records, representing 66 million kilometers of driving across Australia, mainly captured in 2017 and 2018 using over 11,000 unique telematics devices. Each record covers a 30-second period in which various measures, including vehicle speed, acceleration and deceleration were recorded. Vehicle speed was a point measurement (km/hour), whereas acceleration and deceleration were recorded at 1 Hz, but reported as the number of times a fixed g-force threshold was exceeded during the 30-second observational period. Records where a vehicle had not moved at all or where the GPS location was inaccurate were excluded.

Note that the sample of drivers for which telematics data was captured may not be representative of the full driving population. For example, drivers had decided to opt-in for an insurance policy where their driving behavior would be monitored. This resulted in some biases related to sociodemographics and risk characteristics. For example, young drivers were overrepresented, while the potential for reduced insurance premiums could make this product more attractive for safe drivers.

The process of matching telematics data to intersections is independent from the use of satellite imagery, AEs, and clustering, as it only requires the latitude and longitude coordinates of the intersections. It was determined how many vehicles had passed at each of the 898,418 intersections and the recorded measurements of driving behavior were then assigned to the corresponding intersection. The hard acceleration (HA) and hard deceleration (HD) frequencies per intersection were computed as the average number of events per second for a single vehicle. To improve robustness of these computations, only the 25% of intersections with at least 25 observations were analyzed. HA and HD events were defined as exceeding thresholds of 0.15 g and −0.5 g, respectively. The HD threshold is more extreme than the HA threshold, leading to a lower number of recorded events. For example, Papazikou, Quddus, and Thomas (2017) described that 1% of acceleration records were greater than 0.25 g in a large naturalistic driving study, whereas only 0.018% of deceleration records were more extreme than −0.5 g. However, the −0.5 g threshold is a proxy for crashes and near misses. Bagdadi (2013) reported that a threshold of −0.48 g yields a success rate of 76% in detecting near crashes, based on data from the 100-car naturalistic driving study (Dingus et al., 2006). For reference, decelerations exceeding −0.5 g fall in a category described by Boodlal and Chiang (2014) as "very aggressive driving manoeuvres that could result in injury or cause vehicle passengers or cargo that are not securely restrained to be shifted within the vehicle."

After the traffic volume and driving behaviors per intersection were determined, they were assigned to points in the t-SNE embedding. Specifically, as each point in the t-SNE embedding represented a single intersection, the corresponding geographic location was used to make this mapping. Assigning extreme driving events to corresponding intersections in the t-SNE embedding allowed for a data-driven comparison of clusters with similar intersection designs.

## 3 | RESULTS

### 3.1 | Feature extraction

Figure 4 shows the results of first preprocessing several sample images and then passing them through the trained AE (i.e., model inference). This figure indicates that features extracted from the satellite imagery are related to high-level road characteristics concerning the shape, size, and structure of the intersection, while elements less important for road safety disappear.

The bottleneck layer of the AE results in substantial compression of the imagery before regeneration (i.e., about 99.8% compression when taking into account the sparsity constraint). Although our research has found it is possible to regenerate highly accurate images by relaxing the sparsity constraint, capturing mainly road geometry and lane markings as presented here leads to better clustering than also encoding detailed bushes, trees, and other urban features.

### 3.2 | Image clustering

The t-SNE embedding is presented in Figure 5, with each dot representing a single intersection, mapped from the 2,048-dimensional AE representation to a 2-dimensional space. Results of the two validation analyses are presented in the top row of Figure 5. The symbols O, X, T, and # in the figure's legend represent the simplified classes of roundabout, simple four-way, three-way, and complex intersection, respectively. Specifically, the first plot shows one roundabout cluster, a simple four-way intersection cluster, two T-intersection clusters, and a few smaller complex intersection clusters. It is expected that t-SNE does not create a perfect split into these four simplified classes, as the input images contain further details that are also taken into account in the clustering. However, the appearance of intersection groups of similar types based on hold-out information provides a positive initial assessment of the clustering approach. Further, a cluster of signalized intersections appears, even though traffic light information was not supplied to the model during training (i.e., the method is purely image-based), indicating the complexity of the intersection is extracted from imagery. This all provides confidence in the presented framework.

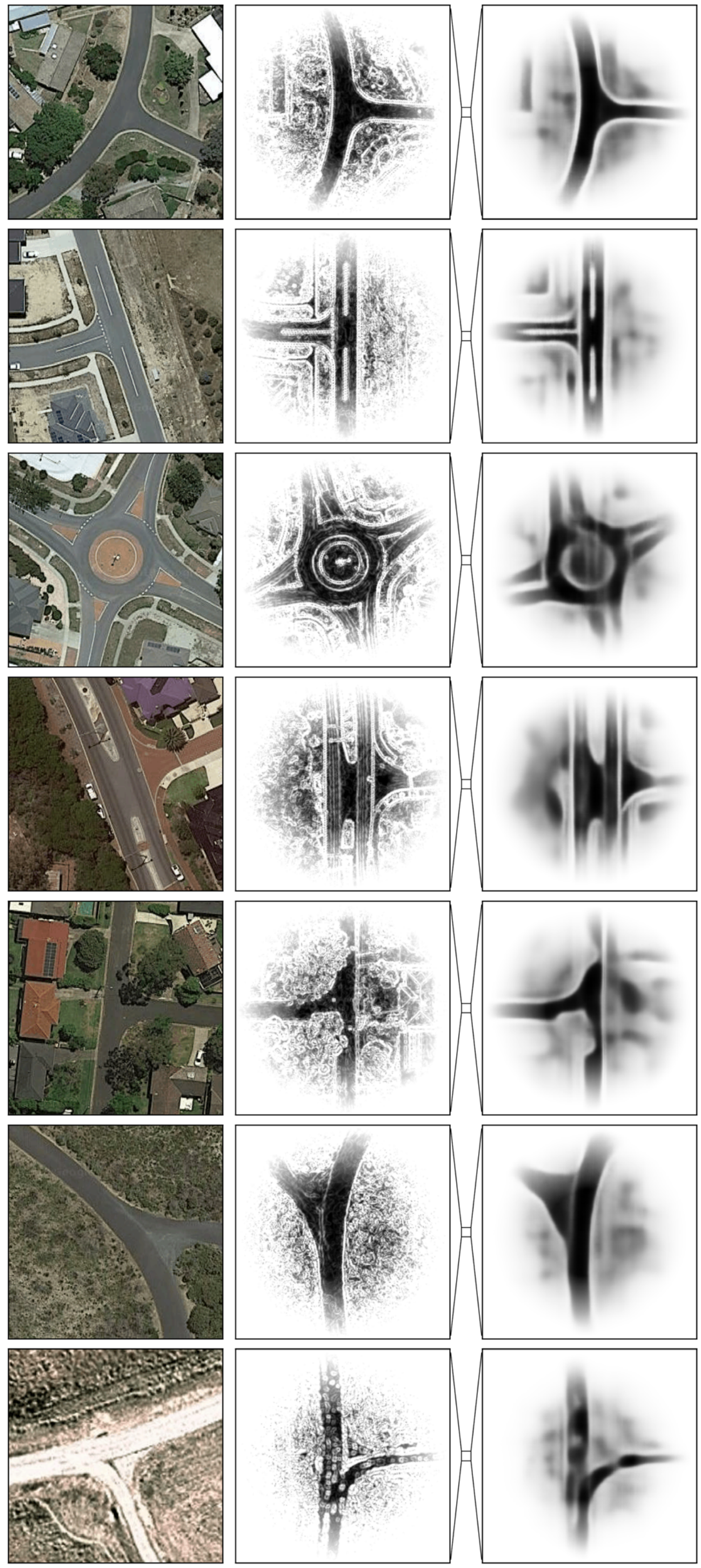

FIGURE 4 Illustration of feature extraction process, showing original (left), preprocessed (middle), and AE regenerated (right) images. The encoded features are retrieved from the AE's bottleneck layer (represented by a small square)

The remaining plots in Figure 5 show the recorded average speed per intersection, traffic volume, and frequencies of HA and HD events. A few outliers were observed in the

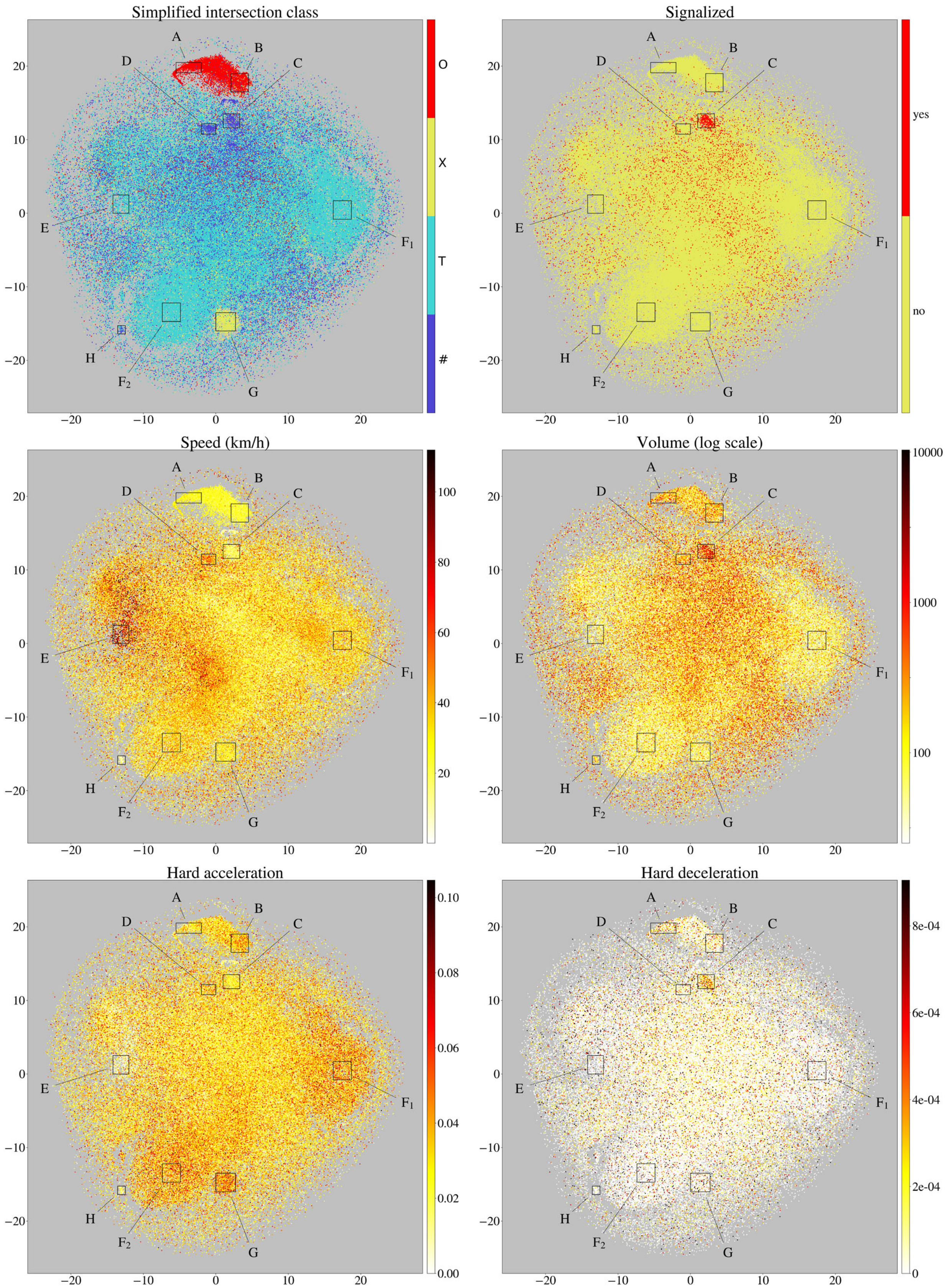

**FIGURE 5** t-SNE clusters with matched OSM and telematics data (best viewed in color)

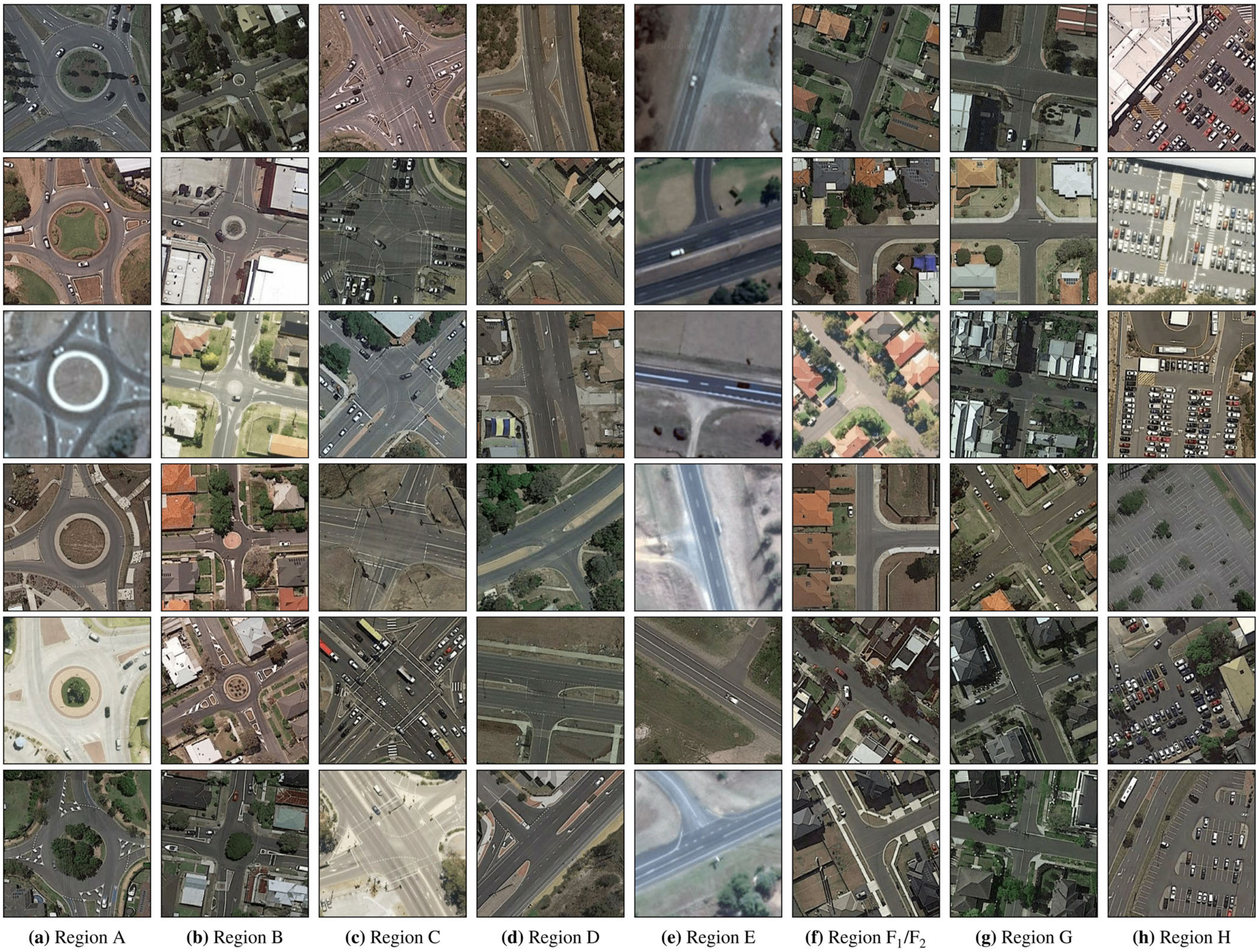

**FIGURE 6** Each column provides sample images for the corresponding region in Figure 5. (a) Region A, (b) Region B, (c) Region C, (d) Region D, (e) Region E, (f) Region $F_1/F_2$, (g) Region G, and (h) Region H

acceleration and deceleration charts, which were removed for plotting purposes. These telematics-based variables all show substantial clustering when plotted on top of the t-SNE embedding and are discussed in detail in the next section.

## 3.3 | Design implications

The regions A–H (see Figure 5) have been manually selected for illustrative purposes. As the t-SNE algorithm captures both the local and global structure of the high-dimensional data, it can be used to explore clusters at several scales (Van der Maaten & Hinton, 2008). Therefore, the position and size of a region can be selected depending on the research question. Samples of intersection designs are provided in Figure 6 for each of the regions selected here. Regions $F_1$ and $F_2$ represent similar designs, as in $F_1$ the preprocessed images have the third leg of the intersection facing left, whereas in $F_2$ this leg faces right.

**TABLE 2** Average recorded driving behavior per intersection in selected regions

| Region | Speed | HA freq | HD freq |
|---|---|---|---|
| A – Large roundabout | 26.3 | 0.027 | 1.6E-04 |
| B – Compact roundabout | 22.9 | 0.036 | 1.1E-04 |
| C – Complex intersection | 19.5 | 0.024 | 2.6E-04 |
| D – Complex T-intersection | 45.3 | 0.033 | 1.1E-04 |
| E – T with access road | 64.6 | 0.016 | 6.1E-05 |
| F – Three-way intersection | 32.4 | 0.038 | 6.8E-05 |
| G – Four-way intersection | 27.6 | 0.043 | 1.2E-04 |
| H – Parking lot | 12.7 | 0.016 | 3.8E-05 |

Furthermore, statistics and box plots are presented in Table 2 and Figure 7, respectively. HA and HD frequencies were measured as the number of events per second. For example, average HD frequencies across intersection types were mostly between 0.0001 and 0.0002 events per second, equivalent to 0.01% to 0.02% of the time (see Table 2). Note that this is the same order of magnitude as mentioned in Papazikou et al. (2017), who reported "values under −0.5 g represent only 0.018% of the total deceleration values." Our research attributes this to different intersection types.

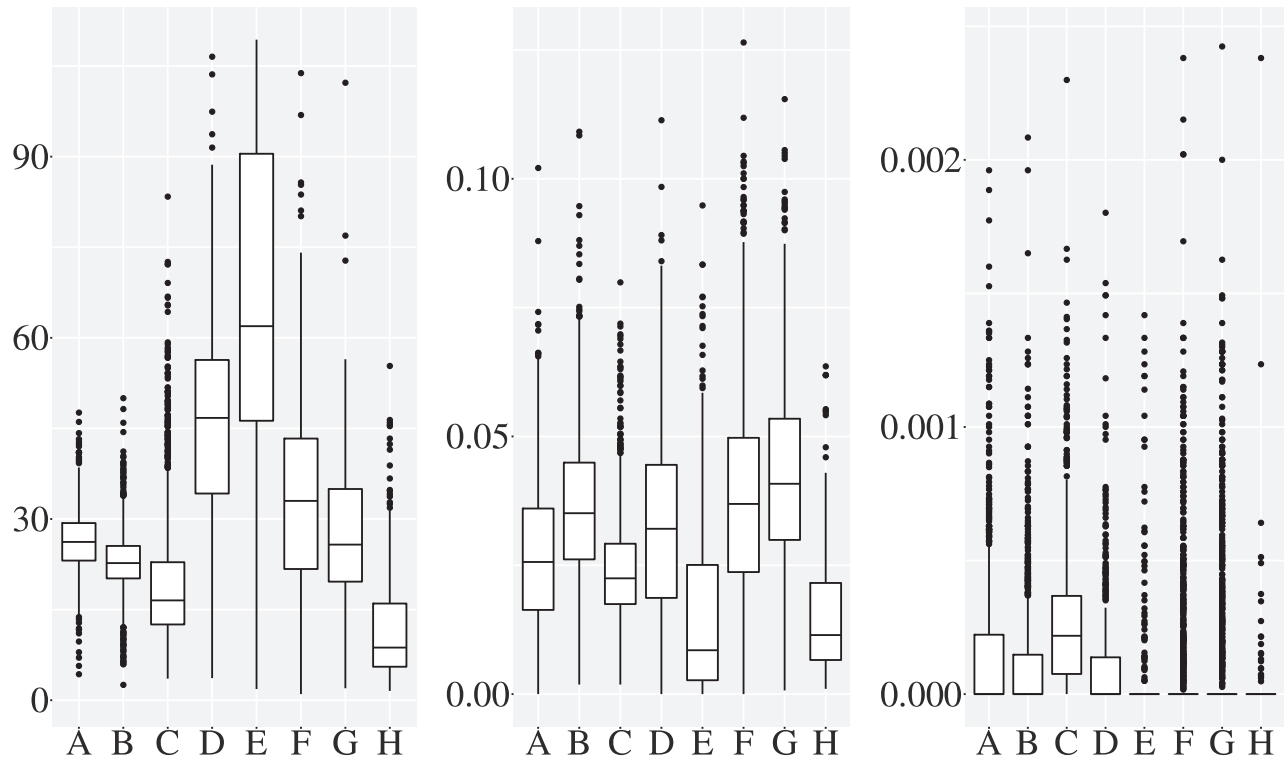

**FIGURE 7** Box plots of data in Table 2. (a) Speed (km/h), (b) HA frequency, and (c) HD frequency

One-way ANOVA tests indicated significant differences in means between regions for speed ($p < 0.001$), acceleration frequencies ($p < 0.001$), and deceleration frequencies ($p < 0.001$). In the following paragraphs, "significant" refers to $p$-values smaller than 0.01 for pairwise comparisons using the Games–Howell post hoc test (i.e., for unequal group variances and without the assumption of normal distributions).

#### 3.3.1 | Roundabouts

The recorded average vehicle speeds on roundabouts (A, B) were significantly lower than on most other crossings (D, E, F, G). In addition, average speed per roundabout did not vary as much as for other designs. Some differences can also be observed within the roundabout cluster. In particular, for compact designs (B) 33% more HA events were observed than for larger designs (A), a significant difference. In contrast, the larger designs showed significantly more HD events.

#### 3.3.2 | Three- and four-way intersections

T-intersections (F) and four-way crossings (G) generally have relatively low traffic volume. However, the latter recorded 12% more HA events: the highest frequency of all clusters (significant for all pairs with G). Three-way intersections (F) recorded comparatively few HD events (significant, except F–D with $p = 0.02$, F–E, and F–H).

The traffic volume at complex intersection cluster (C) is one to two orders of magnitude larger than at the aforementioned clusters. These high-volume, complex intersections are mostly signalized and have the highest frequency of HD events (significant difference for all pairwise comparisons with C). In contrast, designs following the complex T-intersection cluster (D) are generally not signalized, have significantly lower traffic volume and HD events, but significantly higher average speeds and HA events.

### 3.4 | Identifying unsafe designs

Outliers in the deceleration data can be used to identify particularly unsafe designs within each cluster. In this analysis, the average HD frequency in the same region of the t-SNE cluster was used as a benchmark. The following examples (see Figures 8a–c) all had a frequency of HD events (i.e., a proxy for near misses) at least eight times more than average for their category. For each of these examples, at least 175 observations were available of drivers crossing the intersection. It is described where the design features differ from safe designs in the same region, where no HD events were recorded, although causation has not been investigated. The first example (see Figure 8a) is a compact roundabout with several design features requiring additional attention from a driver. In particular, the roundabout has a stop line at one of the roundabout exits for a pedestrian crossing. When a driver has to stop at the line, the roundabout is blocked, forcing circulating cars behind it to stop. Separately, the bicycle lane that ends at the top-right leg could contribute to unsafe situations involving cyclists. Figure 8b shows a very large, multi-lane roundabout with above average traffic volume and speeds; at this location 10.2 times the average HD frequency of region A was observed. For three-way intersections, Figure 8c provides an example of a design outlier with increased HD frequency. The presence of 90° angle parking both near and within the intersection, from which drivers pull out blindly, increases the number of conflict points and potential for unexpected maneuvers.

It is also possible to query the intersection database using specific filters. For example, as intersections in region C are generally signalized, a query was used to identify all intersections in this cluster without traffic lights. Two samples are provided in Figures 8d and e. This could provide opportunities for considering adding traffic lights at selected intersections to improve road safety. Note that detailed design rules exist on when to add traffic lights to

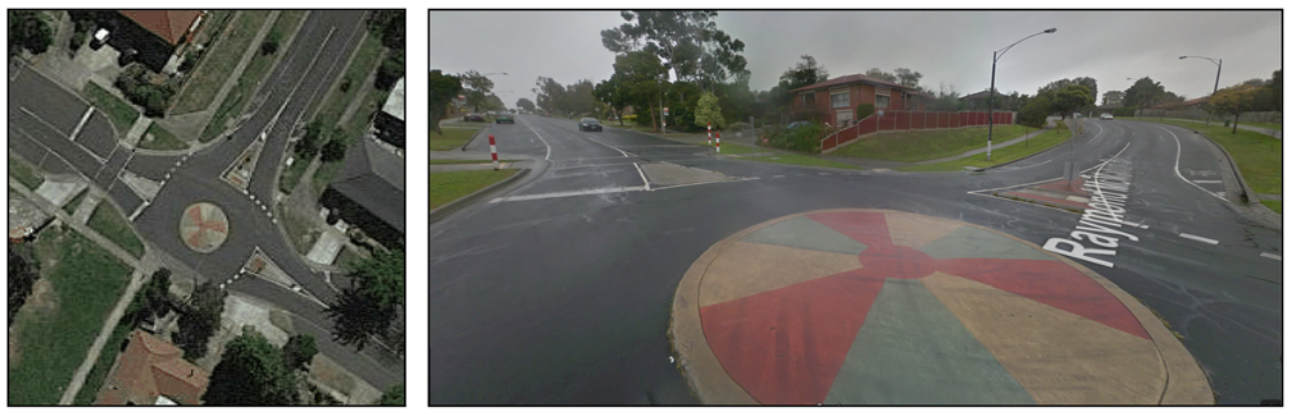

**(a)** Compact roundabout with stop line and ending bicycle lane (region B)

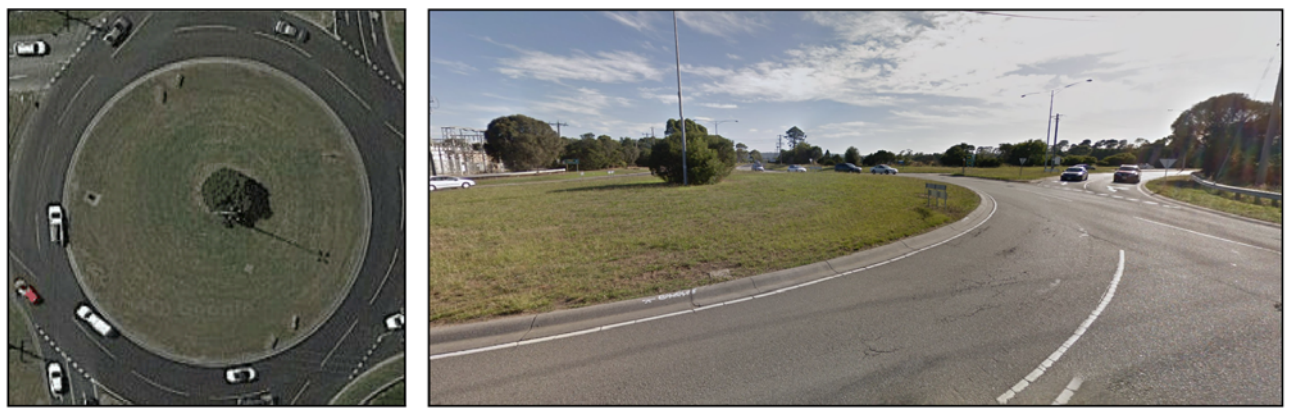

**(b)** Large roundabout with high traffic volume (region A)

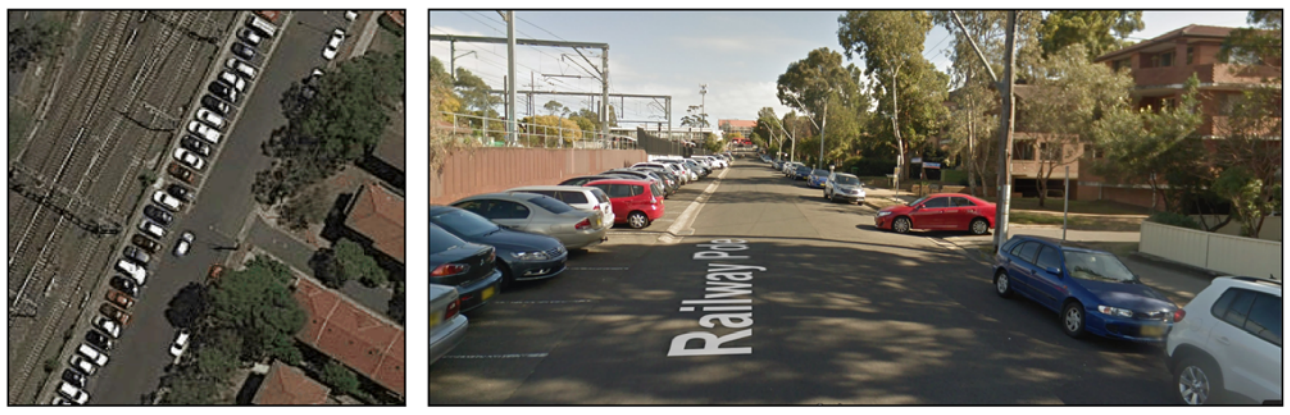

**(c)** Parking spots along T-intersection (region F)

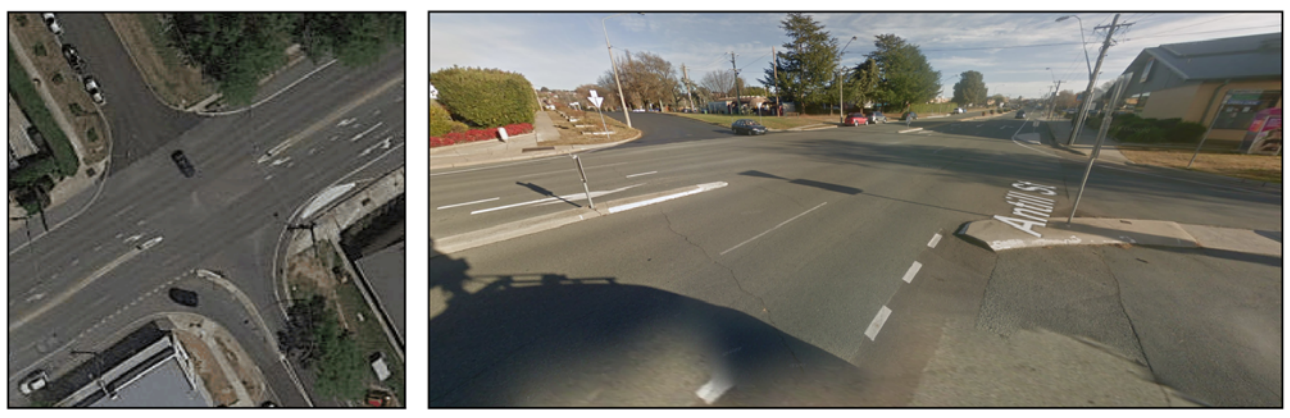

**(d)** Complex intersection without traffic lights (region C)

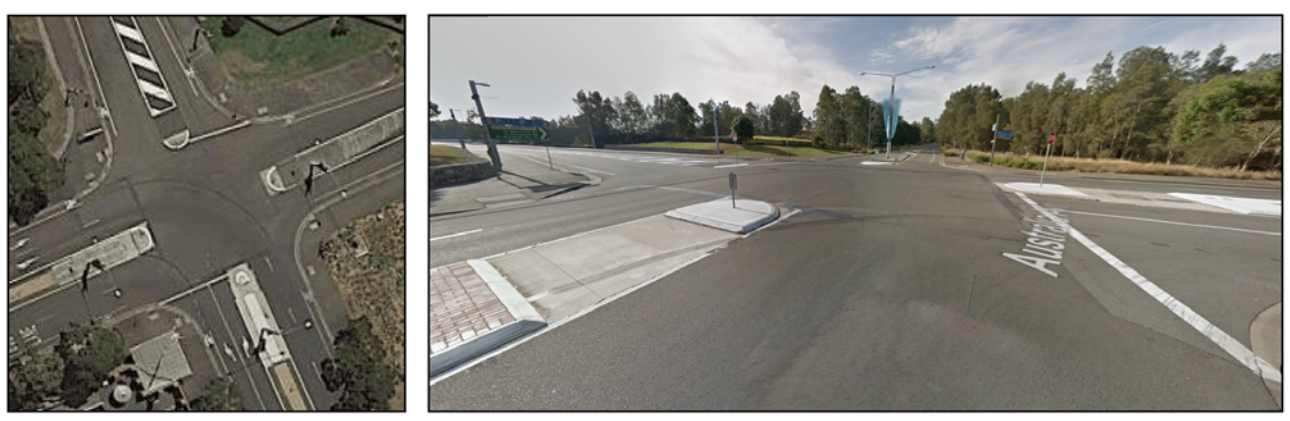

**(e)** Intersection with bus-only leg, no traffic lights (region C)

**FIGURE 8** Google satellite (left) and street view (right) images at selected locations. (a) Compact roundabout with stop line and ending bicycle lane (region B), (b) Large roundabout with high traffic volume (region A), (c) Parking spots along T-intersection (region F), (d) Complex intersection without traffic lights (region C), and (e) Intersection with bus-only leg, no traffic lights (region C)

a specific intersection. In Figure 8e, upon closer inspection, one of the legs is a bus-only leg. Based on the satellite imagery, the AI method expected this to be a signalized intersection.

## 3.5 | Comparison to crash statistics

Results were compared to a database containing 118,082 unique crash events in the State of Victoria between

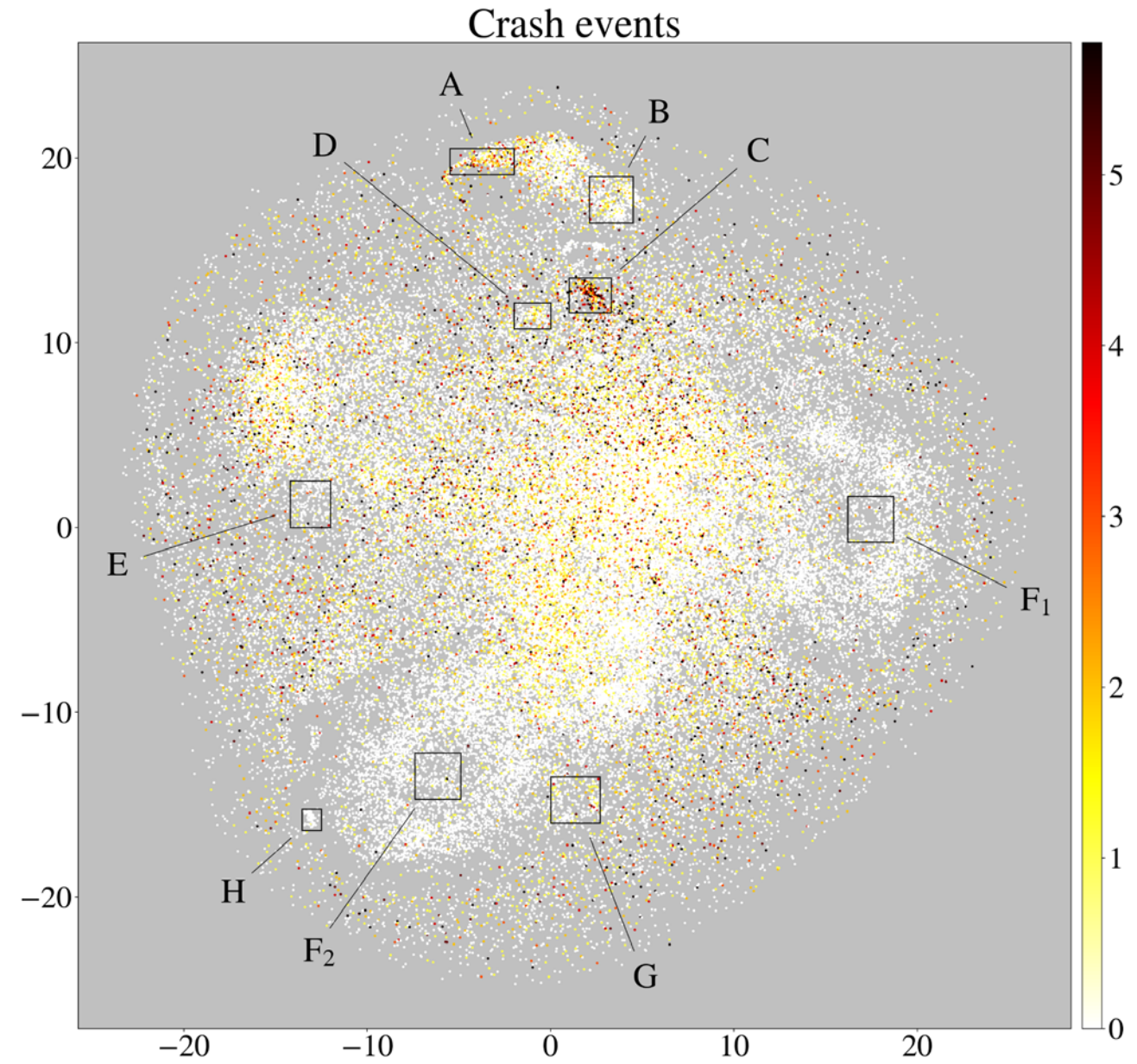

**FIGURE 9** t-SNE clusters with matched crash events

January 2010 and October 2019. Geographic coordinates were available for 73,068 events, and 45,756 of them occurred near intersection locations identified in our research. Figure 9 shows the number of recorded crash events per intersection, plotted using the t-SNE embedding. For better visualization, intersections with more than five crash events have been displayed as black.

For a statistical comparison to these crash statistics, the total number of HD events per intersection were computed, taking into account traffic volume. A correlation of 56% was observed between the number of HD and crash events per intersection. Further, a logistic regression model was fitted with variables volume and HD frequency, and a binary response variable indicating whether a crash had occurred at each intersection. The model achieved an area under the receiver operating characteristic curve (AUC) of 0.86 (Sing, Sander, Beerenwinkel, & Lengauer, 2005). Note that AUC values between 0.8 and 0.9 indicate excellent discrimination (Hosmer, Lemeshow, & Sturdivant, 2013, p. 177).

## 3.6 | Model validation and sensitivity analysis

In this section, the model's robustness and the impact of various assumptions made in this research were explored. First, the data set was split in a 50% training fold and a 50% test fold for validation purposes. The AE model was calibrated using the training fold only (i.e., a much smaller data set than for the main analysis presented earlier). After model calibration, the training and test folds were both processed through the AE to obtain the activations in the bottleneck layer. It was expected that similar features would be extracted from both folds. To visualize any discrepancies, t-SNE was run on the combined results to get a consistent mapping from 2,048-dimensional to 2-dimensional space. Figure 10a illustrates that the AE produces similar results on unseen data.

One of the contributions of this research is the development of specific preprocessing steps to achieve clustering based on features related to road safety. This was an iterative process leading to the final set of preprocessing steps presented in this article. In Figure 10b, some results of the early experiments are presented. The hold-out OSM information was used to illustrate the effects of preprocessing steps on the quality of the embedding. The left plot of Figure 10b shows the results of using satellite imagery directly, without further preprocessing (as in Figure 2a). It is clear that clustering based on raw satellite imagery does not lead to good results. Specifically, by plotting the images on top of this t-SNE embedding (results not presented), it was found that they were simply sorted by the main colors present in the imagery. The use of unrotated imagery (as in Figure 2c) led to smaller clusters based on road orientation, which mainly affected T-intersections.

Figure 10c exemplifies the importance of the sparsity constraint that was added to the loss function (see Equation (2)). Low values of $\beta$ resulted in very accurate image regeneration, by emphasizing reconstruction loss. However, minimizing reconstruction loss alone does not yield optimal results for domain-specific feature extraction. In particular, low values of $\beta$ resulted in clustering based on finer details less important for road safety, now captured in the bottleneck layer (e.g., the specific shape of vegetation). In contrast, strongly enforcing sparsity in the bottleneck layer means the compressed information is not sufficient to accurately capture the type of intersection (right plot of Figure 10c). The selected value for this hyperparameter ($\beta = 0.05$) achieved a balance between these extremes, where some finer details were disregarded while still maintaining sufficient capacity to store important information.

The specification of the AE network architecture (e.g., the number of layers, filter size, and layer depth) has a more subtle impact on results. For example, Figure 10d shows the impact of the size of the bottleneck layer. When the size of the latent space was reduced, the intersection type was still identified well. However, inspecting the AE's regenerated images revealed that most other details had disappeared. On the contrary, when increasing the size of the bottleneck layer, the amount of detail in the regenerated imagery plateaued. As sparsity of this layer is enforced through the loss function, the additional capacity in the latent space is not fully utilized. Because of reduced model complexity, the size of the latent space was set to 2,048.

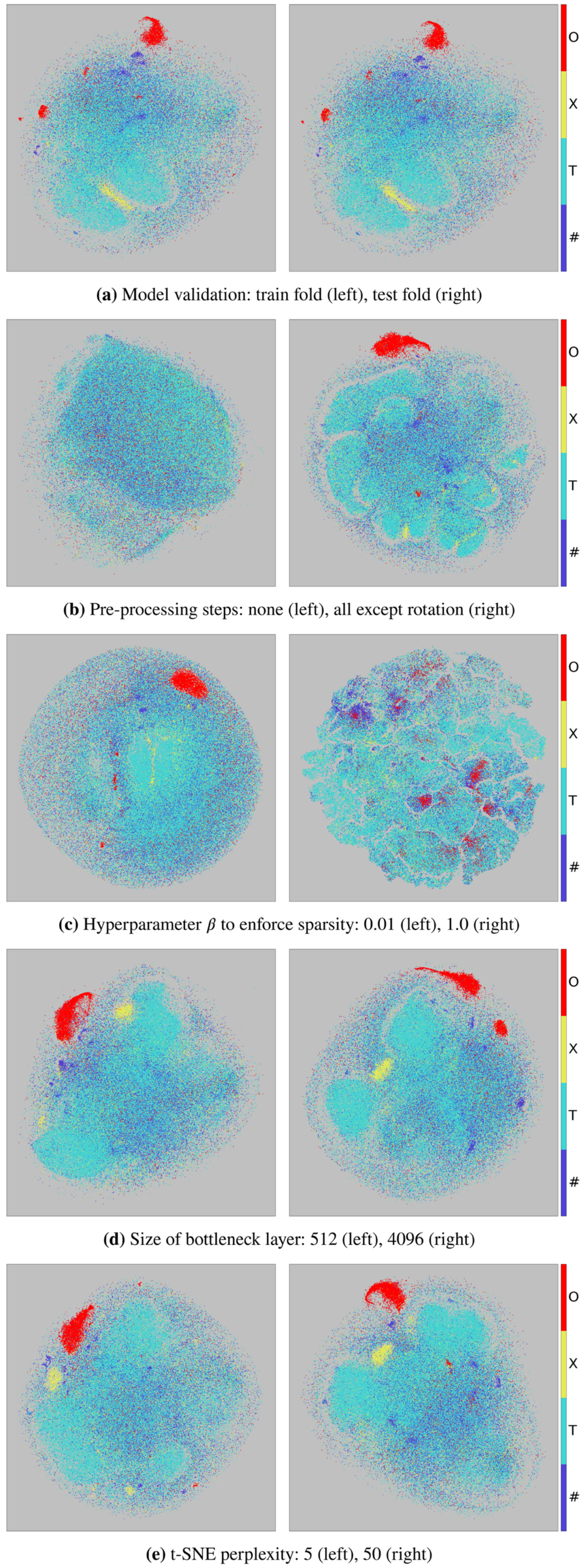

**FIGURE 10** Plots of hold-out information for cross-validation (a) and sensitivity analyses (b–e). (a) Model validation: train fold (left), test fold (right), (b) Preprocessing steps: none (left), all except rotation (right), (c) Hyperparameter $\beta$ to enforce sparsity: 0.01 (left), 1.0 (right), (d) Size of bottleneck layer: 512 (left), 4,096 (right), and (e) t-SNE perplexity: 5 (left), 50 (right)

t-SNE perplexity had a relatively limited impact on the results (Figure 10e), as also noted by developers of the opt-SNE algorithm (Belkina et al., 2019). Values at the higher end of the recommended [5, 50] range resulted in slightly better clustering of complex intersections. In this research, the perplexity parameter was kept at its base setting of 30.

## 4 | DISCUSSION

Our results show that the treatment of satellite imagery using custom preprocessing and deep learning methods, can generate clearly identifiable clusters of intersections that are further associated with differences in driving behavior. A single data source was used for all satellite imagery to ensure consistent data quality. Image blur was only observed for a very small fraction of images, while variations in image brightness were adjusted for using grayscale conversion and the Scharr operator. As the AE will use any information that is still available after preprocessing, the zoom level of a satellite image has a direct impact on feature extraction. The effects of image resolution were not investigated, although high-resolution images may require an increased number of convolutional layers in the AE to extract features at a similar level of abstraction. The sensitivity analyses in Section 3.6 described the impact of several key assumptions on the results. Additional experiments with alternative model configurations did not lead to better clustering, such as a sparsity constraint based on Kullback–Leibler divergence, and pretrained network architectures for the encoding part of the AE.

A limitation of our study is that the OSM data on signalized intersections was not always accurate (i.e., traffic lights observed in Google Street View images were sometimes missing in OSM). Note that this only affects the validation analysis and not the proposed methodology.

In Section 3.3, the selected regions A–H are for illustrative purposes only and different regions may be of interest depending on the specific research question. For example, future research could use smaller, adjacent regions within the embedding to explore nuances in design associated with increased crash risk or dangerous driving behavior, or use the full embedding as the basis for advanced statistical modeling. The embedding could also be used to select geographic locations forming infrastructure baselines for the analysis of dynamic data, such as in-vehicle video footage from naturalistic driving studies. Further, the quality of the embedding could be refined by adding other features extracted from imagery, as identified in previous research. These features include elevation data (Chen, Tang, Zhou, & Cheng, 2019) and lane number or width changes (Bar Hillel, Lerner, Levi, & Raz, 2014), which could influence behavior while approaching the intersection. Further, sight distance could be computed using Geographic Information Systems, while accounting for obstacles like trees or buildings (Castro, Iglesias, Sánchez, & Ambrosio, 2011). By combining these features with the intersection design elements our method extracts from satellite imagery (i.e., before applying the t-SNE algorithm), further improvements could be obtained.

Standard-setting bodies such as AASHTO provide detailed guidelines for the construction of safe, cost-effective intersections with sufficient capacity for the efficient movement of traffic. Big data analyses as presented here enable a data-driven evaluation of intersection safety after the infrastructure has been created. Evaluation using large-scale measurements of actual driving behavior could provide new perspectives on (i) accident black spot identification and (ii) the efficacy of guidelines currently adopted, and has the potential to inform new policies.

## 5 | SUMMARY AND CONCLUSIONS

Our research developed a new methodology to assess the impact of intersection design on road safety, based on computer vision and deep learning. First, all intersection locations in Australia were identified and corresponding satellite imagery was collected. A series of customized preprocessing steps was developed to emphasize features in raw satellite imagery that are related to road safety. High-level features were then extracted from the condensed representation in a deep AE. A loss function was formulated to enforce sparsity in the AE's bottleneck layer, prioritizing feature quality over perfect image reconstruction. The images were clustered using t-SNE based on the extracted features. Finally, driving behavior was explored using recorded telematics data. It was found that HA events (per vehicle) were more frequent at simple four-way than at three-way intersections. T-intersections recorded one of the lowest HD frequencies (a proxy for crash and near-miss events). Further, consistently low average speeds were recorded on roundabouts, contributing to the lower crash severity observed in previous studies. Moreover, several particularly unsafe design features were identified within clusters of similar intersections.

Various validation analyses provide confidence in the proposed methodology, including (i) a comparison to hold-out OSM information, (ii) model evaluation using train and test sets, (iii) sensitivity analyses, and (iv) a comparison to crash statistics. Overall, our method provides a new perspective on design issues that have long been investigated through small-scale case studies using police-reported crash data. In contrast to existing approaches, our research provides the following contributions:

1. Advances in AI methodologies, toolkits, and computing power, allow for the analysis of very large data sets. Rather than the in-depth investigation of a single or few intersections, our study shows how AI enables the investigation of all intersections in a large country.
2. Previous studies used pretrained CNNs and color information to extract features from aerial imagery (e.g., Najjar et al., 2017; Zhang et al., 2018). In contrast, the results of our preprocessing steps show the importance of more domain-specific feature extraction, as a CNN will easily differentiate raw satellite imagery based on features unrelated to road safety. Our approach extracts features including the intersection's type, size, shape, lane markings, and complexity. These features are not easily collected manually and at such a large scale, supporting a computer-aided approach.
3. Rather than using predefined types of intersections (i.e., observer bias), similar intersections are obtained through unsupervised clustering based on actual designs. This allows for the identification of small or large subsets of very similar intersections spread across a large geographic area, taking into account nuances of safety-related features.
4. In-vehicle telematics technology provides measurements of speed, HA and HD behaviors, complementary to crash statistics. By optimizing or eliminating specific design features that have been linked empirically to extreme driving behaviors, countermeasures can be developed to achieve safer intersection designs.

## ACKNOWLEDGMENTS

This research was supported by the ACT Road Safety Fund (grant number 17/8281). Further, the project was undertaken using the LIEF HPC-GPGPU Facility hosted at the University of Melbourne, established with the assistance of LIEF Grant LE170100200. J.T. is supported by an Australian Research Council Discovery Early Career Researcher Award (grant number DE180101411). M.S. is supported by a National Health and Medical Research Council (Australia) Fellowship (grant number APP1136250). J.G. is supported by the Opening Foundation of the Key Laboratory of Road and Traffic Engineering of the Ministry of Education (China). The authors would like to acknowledge the 10 anonymous reviewers for their valuable feedback, which helped improve the quality of the original article.

## SUPPORTING INFORMATION

The following supporting information is available on Zenodo: Data set S1 (https://doi.org/10.5281/zenodo.2564253)

**Data set S1.** This data set contains 898,418 preprocessed images (see Figure 2f) of all identified intersections in Australia, including their geographic coordinates (Wijnands, 2020).